\documentclass{article}
\usepackage{iclr2027_conference,times}

\usepackage{amsmath,amsfonts,bm}

\def\eqref#1{equation~\ref{#1}}

\def\1{\bm{1}}

\DeclareMathAlphabet{\mathsfit}{\encodingdefault}{\sfdefault}{m}{sl}
\SetMathAlphabet{\mathsfit}{bold}{\encodingdefault}{\sfdefault}{bx}{n}

\usepackage{hyperref}
\usepackage{xurl}
\hypersetup{
  hidelinks,
  pdftitle={DeFA: Dependency-Guided Failure Attribution for LLM Agents},
  pdfauthor={Bo Deng, Xinlei Zheng, Yi Wei, Kang Zhou, Chongyang Tao, Renzhao Liang, Xuanren Chen, Lifan Guo, Chi Zhang}
}
\usepackage{booktabs}
\usepackage{array}
\usepackage{flafter}
\usepackage{graphicx}
\usepackage[breakable]{tcolorbox}
\usepackage[utf8]{inputenc}
\usepackage{fvextra}

\AddToHook{env/table/begin}{\setlength{\belowcaptionskip}{\baselineskip}}

\newcommand{\ARXIVBODYONLY}{}

\title{DeFA: Dependency-Guided Failure Attribution for LLM Agents}

\author{
  \textbf{\href{https://openreview.net/profile?id=~Bo_Deng2}{Bo Deng}}$^{1,2}$ \quad
  \textbf{\href{https://openreview.net/profile?id=~Xinlei_Zheng1}{Xinlei Zheng}}$^{2}$ \quad
  \textbf{\href{https://openreview.net/profile?id=~Yi_Wei10}{Yi Wei}}$^{2}$ \\
  \textbf{\href{https://openreview.net/profile?id=~Kang_Zhou8}{Kang Zhou}}$^{1}$ \quad
  \textbf{\href{https://openreview.net/profile?id=~Chongyang_Tao1}{Chongyang Tao}}$^{2}$ \quad
  \textbf{\href{https://openreview.net/profile?id=~Renzhao_Liang2}{Renzhao Liang}}$^{2}$ \\
  \textbf{\href{https://openreview.net/profile?id=~XuanrenChen1}{Xuanren Chen}}$^{2}$ \quad
  \textbf{\href{https://openreview.net/profile?id=~Lifan_Guo2}{Lifan Guo}}$^{1}$ \quad
  \textbf{\href{https://openreview.net/profile?id=~Chi_Zhang103}{Chi Zhang}}$^{1}$ \\[0.6em]
  {\normalfont $^1$Qwen DianJin Team, Alibaba Cloud Computing} \\
  {\normalfont $^2$\href{https://www.buaa.edu.cn/}{Beihang University}}
}

\iclrfinalcopy
\begin{document}

\maketitle
\lhead{}

\begin{abstract}
Errors in LLM agent executions and their visible consequences can be separated by many steps, making decisive-error localization a matter of understanding both step content and step dependencies. We introduce DeFA, a dependency-guided framework for agent failure attribution. DeFA first combines protocol relations and semantic dependencies into an event dependency graph spanning the trajectory. It then identifies events that may violate task requirements and traces their sources and subsequent effects to construct a failure propagation graph. Finally, DeFA uses step evidence and the steps' roles in failure propagation to identify the decisive error, responsible agent, and error category. To support long trajectories, DeFA partitions executions into segments and combines the current segment's detailed content with summaries of the other segments, giving local diagnosis access to global execution context. Across Who\&When and the Who\&When Pro text subset, DeFA achieves the highest responsible-agent and exact step accuracy with all evaluated backbones, and the highest failure-mode accuracy among taxonomy-aligned methods on Pro. Further experiments on image and video trajectories demonstrate its applicability to multimodal failure attribution. Ablations support the contributions of segmentation, the event dependency graph, and the failure propagation graph. Using DeFA's diagnostic feedback for skill evolution in Trace2Skill improves downstream task accuracy by 6--15 percentage points over the native pipeline, showing that the diagnoses can also support agent improvement on subsequent tasks.\ifdefined\ARXIVBODYONLY\else\ Code is available at \url{https://anonymous.4open.science/r/defa-code-FA22/}.\fi

\end{abstract}

\section{Introduction}
\label{sec:introduction}

When an LLM agent fails a task, the final outcome rarely reveals which earlier decision was responsible. Agents interleave reasoning, tool use, and communication~\citep{yao2023react,wu2024autogen}, producing trajectories in which an error and its visible consequences may be separated by many steps. Failure attribution identifies the responsible agent and the decisive error step~\citep{zhang2025whowhen,liu2026whowhenpro}, providing specific evidence for debugging and subsequent improvement.

The difficulty lies in tracing how errors arise and propagate through execution to identify the decisive error behind downstream symptoms. Suppose an agent misreads a retrieved value, another agent uses it in a calculation, and the final answer repeats the incorrect result. The final answer reveals the failure, but the earlier interpretation may be the decisive error. Empirical studies likewise examine failure onset and propagation in agent trajectories~\citep{zhao2026failureprocess}. Attribution must therefore examine what information a step received, how it used that information, and what happened afterward. This becomes harder when relevant evidence is distributed across a long trace~\citep{liu2024lostmiddle}.

Existing attribution frameworks use global judgments, sequential checks, or binary search~\citep{zhang2025whowhen}, as well as constraint validation, multi-agent investigation, and hierarchical diagnosis~\citep{barke2026agentrx,li2026errorprobe,wang2026cockpithat}. Our focus is to connect suspicious events with the execution that made them consequential. A local deviation must be examined together with its upstream sources and the later actions that adopted, amplified, or corrected its effects. Reconstructing these relations brings distributed evidence together to explain how the execution failed.

We introduce \textbf{DeFA}, a framework for \textbf{dependency-guided failure attribution} from recorded trajectories. DeFA first combines protocol relations and semantic dependencies into an event dependency graph spanning the trajectory. Building on this structure, it identifies events that may violate task requirements and traces their sources and subsequent effects to construct a failure propagation graph. It then combines trajectory content and propagation relations to identify the decisive error step, responsible agent, and error category. For long trajectories, DeFA uses segmented analysis, combining the current segment's detailed content with summaries of the other segments so that local diagnosis can use both concrete step evidence and global execution context.

Against seven baselines, DeFA leads in responsible-agent and exact step accuracy on Who\&When with all three backbones, improving exact step by 7.6--9.8 percentage points. On 300 Pro text trajectories, it also achieves the highest responsible-agent and exact step accuracy with all three backbones, as well as the highest failure-mode accuracy among taxonomy-aligned methods. On 100 multimodal trajectories, it improves exact step accuracy by 6.0--7.0 points across both backbones. Ablations show that segmentation, the event dependency graph, and the failure propagation graph each contribute to attribution accuracy. Removing propagation relations lowers exact step accuracy even when the adjudication set is held fixed, supporting their role in identifying the decisive error. Downstream experiments integrate DeFA's diagnostic feedback into Trace2Skill, improving test accuracy over the native pipeline by 6--15 percentage points across human-written and model-generated initial skills and demonstrating its utility for agent improvement on subsequent tasks.

Our contributions are:
\begin{itemize}
    \item We introduce a failure attribution framework that combines protocol relations and semantic dependencies into an event dependency graph to identify suspicious events, trace their contribution to task failure, and locate the decisive error and responsible agent.
    \item We develop a failure propagation graph that connects suspicious events with their upstream sources and downstream effects, tracing how errors are introduced, transmitted, and adopted to support decisive-error attribution.
    \item We demonstrate gains in precise attribution across text and multimodal evaluations, and show that DeFA's diagnostic feedback improves downstream skill evolution by 6--15 percentage points over native Trace2Skill.
\end{itemize}

\section{Related Work}
\label{sec:related-work}

\paragraph{Execution trajectory evaluation and analysis.}
Agent evaluation includes response judging~\citep{zheng2023judging}, intermediate work and competency assessment through Agent-as-a-Judge and AgentDiagnose~\citep{zhuge2025agentjudge,ou2025agentdiagnose}, and reasoning correctness and error localization through process supervision and ProcessBench~\citep{lightman2024verify,zheng2025processbench}. MAST characterizes design, coordination, and verification failures~\citep{cemri2025mast}; ToolEmu evaluates risk in emulated tool environments~\citep{ruan2024toolemu}; and \citet{zhao2026failureprocess} study coding-agent failure development and recovery.

\paragraph{Failure diagnosis and attribution frameworks.}
Who\&When formalizes responsible-agent and decisive-step attribution with global, sequential, and binary-search judges~\citep{zhang2025whowhen}. TRAIL supplies a taxonomy and human-annotated diagnostic traces~\citep{deshpande2025trail}; Who\&When Pro uses controlled error injection across modalities~\citep{liu2026whowhenpro}.

AgentRx supplies validated constraint violations to an attribution judge~\citep{barke2026agentrx}, VerifyMAS tests taxonomy-derived hypotheses~\citep{qiao2026verifymas}, and DeepDebug combines global reading, focused investigation, and cross-examination~\citep{zhu2026agentdebugx}. ErrorProbe detects local anomalies using a failure taxonomy and traces backward from observed symptoms to focus the investigation. Specialized agents then propose, investigate, and verify error hypotheses using executable evidence~\citep{li2026errorprobe}. CockpitHAT performs hierarchical diagnosis for embodied multi-agent cockpits, integrating dialogue and other execution evidence with a panel of analysts. Its consensus procedure weighs analyst confidence and emphasizes failures with greater safety consequences~\citep{wang2026cockpithat}. EDGE studies attribution of multiple error categories within a trajectory, combining a two-stage LLM judge for detection with counterfactual rollout analysis for interpreting errors and potential repairs~\citep{hou2026edge}.

Additional executions provide another source of evidence: AgenTracer constructs training supervision through counterfactual replay and fault injection~\citep{zhang2026agentracer}, FAMAS estimates suspiciousness from variation across repeated executions~\citep{ge2025famas}, and REFLECT and CausalFlow assess responsibility through patched replay or step-level interventions~\citep{lin2026reflect,bonagiri2026causalflow}.

\paragraph{Diagnostic feedback for agent improvement.}
Self-Refine iteratively revises outputs using feedback~\citep{madaan2023selfrefine}; Reflexion carries verbal feedback into later trials~\citep{shinn2023reflexion}. Trace2Skill consolidates trajectory lessons into reusable skills~\citep{ni2026trace2skill}, including for SpreadsheetBench tasks~\citep{ma2024spreadsheetbench}. We evaluate DeFA's diagnostic feedback within this skill evolution process.

\section{Method}
\label{sec:method}

\begin{figure}[!htbp]
    \centering
    \includegraphics[width=\textwidth]{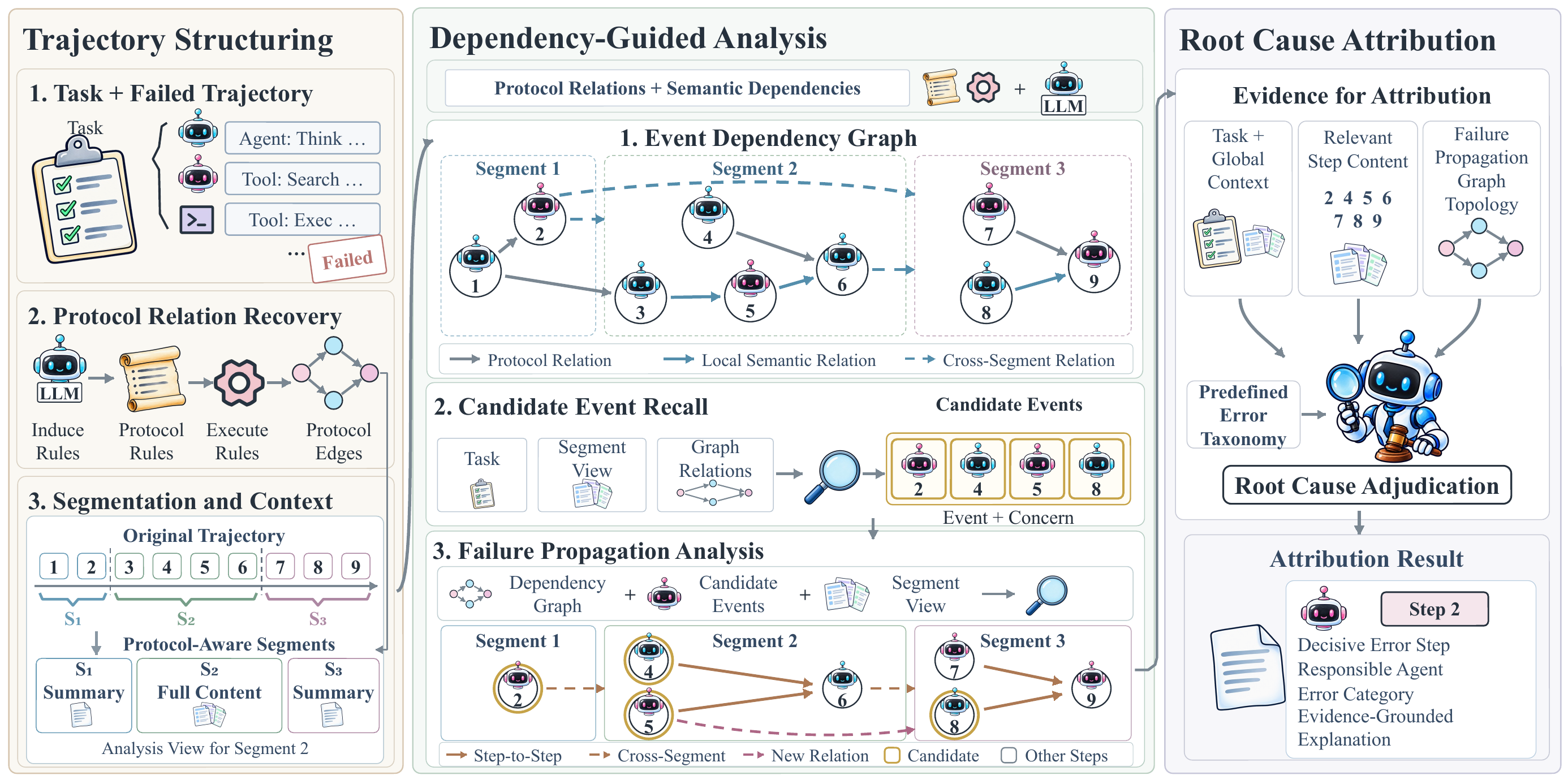}
    \caption{Overview of DeFA's dependency-guided failure attribution pipeline.}
    \label{fig:method-overview}
\end{figure}

DeFA constructs an event dependency graph to guide candidate recall, then traces candidate sources and effects in a failure propagation graph. Final adjudication compares the resulting steps and evidence to identify the decisive error. Figure~\ref{fig:method-overview} summarizes the pipeline\ifdefined\ARXIVBODYONLY.\else; Appendix~\ref{app:method-details} provides implementation details.\fi

\subsection{Problem formulation}
\label{sec:problem-formulation}

Following Who\&When and Who\&When Pro~\citep{zhang2025whowhen,liu2026whowhenpro}, given task $x$ and failed trajectory $\tau=(s_1,\ldots,s_T)$, DeFA predicts the \emph{decisive error step} $\hat{t}$: the earliest erroneous step whose correction would make execution succeed. Its responsible agent is $\hat{a}=a_{\hat{t}}$, where $a_t$ is the actor recorded at $s_t$. The error category $\hat{y}\in\mathcal{Y}$ follows the Pro taxonomy, including hallucination and tool parameter/invocation error\ifdefined\ARXIVBODYONLY.\else\ (Appendix~\ref{app:evaluation-details}).\fi

\subsection{Event dependency graph construction}
\label{sec:trajectory-structuring}
\label{sec:event-graph}

The event dependency graph represents each step as a node in $V=\{1,\ldots,T\}$ and combines protocol relations with semantic dependencies. It records how steps pass and use instructions, results, and state.

Protocol relations capture interactions defined by the agent framework's conventions, such as task delegation and its execution, or tool requests and their results. Semantic dependencies capture how later steps refer to or use earlier content, such as adopting a previous conclusion or using a retrieved value in a calculation.

\paragraph{Protocol relation recovery.}
DeFA first compresses oversized step content into a normalized trajectory $\tilde{\tau}$, preserving step identities and actors under instructions to retain facts, provenance, and action status. An LLM then induces protocol rules from message roles, formats, and interaction patterns. Each rule specifies source and target matching conditions, a pairing procedure, and the resulting relation. A deterministic interpreter scans $\tilde{\tau}$ to form the \emph{protocol edge set} $E_{\mathrm{p}}$: each edge $(i,j)$ links a source step to a matched later step, such as a tool request to its result or a delegation to its execution. The interpreter supports next-match, all-match, FIFO, and LIFO pairing to accommodate different interaction patterns\ifdefined\ARXIVBODYONLY.\else\ (Appendix~\ref{app:protocol-edges}).\fi\ These edges guide segmentation and complement semantic dependencies.

\paragraph{Segmentation and context.}
DeFA forms $M$ consecutive segments $\mathcal{S}=(S_1,\ldots,S_M)$ under step and content budgets, adjusting boundaries using $E_{\mathrm{p}}$ to preserve related interactions where possible. For multiple segments, it generates a narrative summary $u_j$ of each. The analysis view $\mathcal{V}_j$ combines the full normalized content of $S_j$ with the other summaries:
\begin{equation}
    \mathcal{V}_j=\big(u_1,\ldots,u_{j-1},S_j,u_{j+1},\ldots,u_M\big).
\end{equation}
This view limits context length while preserving local detail and broader execution: earlier summaries describe information sources, and later ones show whether information was used, corrected, or abandoned. Semantic dependency recovery, candidate recall, and failure propagation analysis each use this view per segment, retaining original step indices. Single-segment trajectories use the complete normalized trajectory directly.

\paragraph{Semantic dependency recovery.}
For each $S_j$, the model uses the task, $\mathcal{V}_j$, and relevant protocol edges to identify direct uses of earlier content, instructions, results, or state. Each semantic edge explains what its source supplied and how its target used it, such as a calculation using a value from an earlier observation. Local edges connect two steps in the detailed segment and merge into $E_{\mathrm{s}}^{\mathrm{local}}$. Cross-segment edges connect a step in $S_j$ to a later segment $S_k$ ($k>j$) whose summary indicates continued use of its information; these merge into $E_{\mathrm{s}}^{\mathrm{cross}}$. All edges point forward in execution order, forming
\begin{equation}
    G_{\mathrm{c}}=\big(V,\mathcal{S},E_{\mathrm{p}}\cup E_{\mathrm{s}}^{\mathrm{local}},
    E_{\mathrm{s}}^{\mathrm{cross}}\big).
\end{equation}

\subsection{Candidate event recall}
\label{sec:candidate-recall}

For each $S_j$, the recall model $R$ examines every step using task requirements, $\mathcal{V}_j$, and event dependency graph relations $G_{\mathrm{c}}^{(j)}$: step-to-step edges with an endpoint in $S_j$ and cross-segment edges with a source in $S_j$ or a target of $S_j$. It returns candidate records $\mathcal{Q}_j$, merged into $\mathcal{Q}$ with step set $C$:
\begin{equation}
    \mathcal{Q}_j = R\big(x,\mathcal{V}_j,G_{\mathrm{c}}^{(j)}\big),
    \qquad \mathcal{Q} = \bigcup_{j=1}^{M}\mathcal{Q}_j,
    \qquad C=\operatorname{Steps}(\mathcal{Q}).
\end{equation}
Each record retains the original step index, describes the step's action or output, and explains its potential connection to failure using task requirements and trajectory evidence, such as information inconsistent with task constraints or an omitted necessary operation.

\subsection{Failure propagation analysis}
\label{sec:failure-graph}

Using $x$, $\mathcal{V}_j$, $\mathcal{Q}_j$, and $G_{\mathrm{c}}^{(j)}$, DeFA traces candidates' upstream sources and how later execution continued, amplified, or corrected their effects. The model selects relevant edges from the event dependency graph and infers additional propagation edges from trajectory evidence. Each edge explains how source information or results were used and affected the task outcome. Step-to-step edges form $E_{\mathrm{f}}^{\mathrm{exact}}$; edges targeting later segments form $E_{\mathrm{f}}^{\mathrm{cross}}$. Merging segment outputs yields the failure propagation graph $G_{\mathrm{f}}$, whose step set includes candidates, exact edge endpoints, and cross-segment source steps:
\begin{equation}
\begin{aligned}
    G_{\mathrm{f}} &= \big(V_{\mathrm{f}},\mathcal{S},E_{\mathrm{f}}^{\mathrm{exact}},
    E_{\mathrm{f}}^{\mathrm{cross}}\big),\\
    V_{\mathrm{f}} &= C\cup\operatorname{Endpoints}_{V}(E_{\mathrm{f}}^{\mathrm{exact}})
    \cup\operatorname{Sources}_{V}(E_{\mathrm{f}}^{\mathrm{cross}}).
\end{aligned}
\end{equation}
Thus, steps discovered through propagation relations join the initial candidates in $V_{\mathrm{f}}$ for final attribution.

\subsection{Root cause attribution}
\label{sec:root-adjudication}

An LLM adjudicator $J$ selects the decisive error from $V_{\mathrm{f}}$ using the task, failure propagation graph, and trajectory evidence view $\mathcal{Z}$. For a single segment, $\mathcal{Z}$ is the complete normalized trajectory; otherwise, it contains all segment summaries in execution order and the full normalized content of each step in $V_{\mathrm{f}}$.

After identifying why the task failed, the adjudicator uses step content and propagation relations to compare plausible errors and their effects on the outcome. For each plausible root cause, it assesses whether a correct alternative was feasible at that point and whether substituting it, with later execution continuing from the corrected state, would turn failure into success. When several steps qualify, it compares how directly each explains the failure and the supporting evidence. It selects $\hat{t}$ and classifies that step's own error under $\mathcal{Y}$:
\begin{equation}
    (\hat{t},\hat{y})=J\big(x,\mathcal{Z},G_{\mathrm{f}};\mathcal{Y}\big).
\end{equation}
Here, $\hat{t}\in V_{\mathrm{f}}$, $\hat{y}\in\mathcal{Y}$, and $\hat{a}=a_{\hat{t}}$. DeFA returns these predictions with an attribution explanation grounded in trajectory evidence.

\paragraph{Illustrative attribution.}
\ifdefined\ARXIVBODYONLY In an illustrative Who\&When example,\else In the Who\&When example in Appendix~\ref{app:worked-example},\fi\ WebSurfer retrieves a film-ranking table at step 12, Assistant uses it to compare two rankings at step 16, and step 19 outputs the resulting count. The failure propagation graph connects these steps as $12\to16\to19$. Comparing the retrieved table with Assistant's interpretation reveals that the table contains the needed entry, but Assistant omits it from the comparison. The missing film is Demon Slayer, whose inclusion changes the answer from five shared films to the correct count of six. DeFA attributes the failure to step 16, matching the benchmark annotation. The connected evidence identifies where the available information was used incorrectly.

Specifically, the retrieved domestic ranking places Demon Slayer tenth and Wonder Woman 1984 immediately below it. Assistant includes the latter in the domestic top ten, creating the mismatch used in the overlap calculation.

\section{Experiments}
\label{sec:experiments}

\subsection{Experimental setup}
\label{sec:experimental-setup}

\paragraph{Benchmarks.}
We evaluate on Who\&When~\citep{zhang2025whowhen} and Who\&When Pro~\citep{liu2026whowhenpro}. Who\&When contains 184 annotated failure trajectories from algorithm-generated and hand-crafted multi-agent systems; we pool both subsets for evaluation. From Pro, we select 300 text trajectories spanning eight single- and multi-agent frameworks, stratified by framework and failure-mode group with additional coverage constraints on trajectory length and error position. All methods use the same evaluation trajectories.\ifdefined\ARXIVBODYONLY\else\ Sampling details are provided in Appendix~\ref{app:evaluation-details}.\fi

\paragraph{Baselines and backbones.}
We compare DeFA with seven baselines: All-at-Once, Step-by-Step, and Binary Search from Who\&When~\citep{zhang2025whowhen}; AgentRx~\citep{barke2026agentrx}; DeepDebug, the diagnostic component of AgentDebugX~\citep{zhu2026agentdebugx}; ErrorProbe~\citep{li2026errorprobe}; and CockpitHAT~\citep{wang2026cockpithat}. We evaluate every method with GLM-5.2, DeepSeek-V4-Pro, and Qwen3.8-Max on both text benchmarks. For each backbone, all methods use the same model for every LLM invocation, with temperature set to 0 and greedy decoding.

\paragraph{Metrics and scoring.}
We evaluate attribution along three dimensions: responsible agent, error step, and failure mode.
\textbf{Agent} measures responsible-agent accuracy on trajectories from multi-agent frameworks. It uses all 184 Who\&When trajectories and, following Pro's official Who evaluation protocol, 89 Pro text trajectories, excluding Smolagents, ALFAgent, and MathChat.
\textbf{Step} measures exact error-step localization accuracy; Step~$\pm1$ and Step~$\pm3$ allow deviations of up to one and three steps, respectively. These metrics use all 184 Who\&When and all 300 Pro text trajectories.
\textbf{Mode} measures failure-mode classification accuracy on all 300 Pro text trajectories. We compare the three Who\&When judges and DeFA; the other methods do not adopt Pro's failure-mode taxonomy and are therefore excluded from this comparison, with dashes shown in the tables.

We score the final prediction for each trajectory, evaluating each metric independently. Some Who\&When Pro trajectories have multiple acceptable attribution answers. Following the official scoring rules, a prediction is correct on a metric if it matches any accepted answer on that axis. All methods' step predictions are mapped to official coordinates before scoring\ifdefined\ARXIVBODYONLY.\else; adaptation details are provided in Appendix~\ref{app:input-adaptation}.\fi

Each accuracy is the number of correctly predicted trajectories divided by the total number evaluated for that metric.
For Step~$\pm k$ ($\mathrm{S}_{\pm k}$ in tables, $k\in\{1,3\}$), let $\hat{t}_i$ be the predicted step position and $T_i$ the set of error positions accepted by the official scoring rules. A prediction is correct if it lies within $k$ steps of any accepted position, giving the correctness indicator:
\begin{equation}
    c_i^{(\pm k)}=\mathbf{1}\!\left[\min_{t\in T_i}|\hat{t}_i-t|\le k\right].
\end{equation}
Tables bold the best results and underline the second-best distinct results within each backbone, with ties marked alike.

\begin{table}[!htb]
\centering
\caption{Failure attribution accuracy on Who\&When (\%).}
\label{tab:whowhen-main}
\small
\setlength{\tabcolsep}{2.5pt}
\renewcommand{\arraystretch}{1.0}
\begin{tabular*}{\textwidth}{@{\extracolsep{\fill}}l*{12}{r}@{}}
\toprule
& \multicolumn{4}{c}{GLM-5.2} & \multicolumn{4}{c}{DeepSeek-V4-Pro} & \multicolumn{4}{c}{Qwen3.8-Max} \\
\cmidrule(lr){2-5}\cmidrule(lr){6-9}\cmidrule(lr){10-13}
Method & Agent & Step & $\mathrm{S}_{\pm1}$ & $\mathrm{S}_{\pm3}$ & Agent & Step & $\mathrm{S}_{\pm1}$ & $\mathrm{S}_{\pm3}$ & Agent & Step & $\mathrm{S}_{\pm1}$ & $\mathrm{S}_{\pm3}$ \\
\midrule
All-at-Once & 62.5 & 17.9 & 46.7 & 68.5 & \underline{62.5} & 14.7 & 46.2 & \textbf{72.8} & 62.0 & 15.8 & 46.7 & 60.9 \\
Step-by-Step & 46.7 & 27.2 & 41.3 & 60.9 & 45.7 & 22.3 & 41.8 & 60.9 & 41.3 & 23.4 & 41.8 & 59.2 \\
Binary Search & 54.9 & 31.5 & 44.6 & 60.9 & 60.9 & 34.8 & 46.7 & 64.7 & \underline{64.1} & 37.5 & 49.5 & 63.0 \\
AgentRx & 59.8 & 38.6 & \underline{51.6} & 67.9 & 57.1 & 35.9 & 50.0 & 65.8 & 49.5 & 39.7 & 46.2 & 64.7 \\
DeepDebug & 58.7 & 38.6 & 50.0 & \underline{69.0} & 58.7 & \underline{39.1} & \underline{50.5} & 67.9 & 55.4 & 36.4 & \underline{50.0} & \textbf{70.1} \\
ErrorProbe & 59.8 & \underline{43.5} & \underline{51.6} & 65.2 & 57.6 & 36.4 & 44.0 & 61.4 & 59.8 & 39.1 & 47.3 & 65.2 \\
CockpitHAT & \underline{63.6} & 42.9 & \underline{51.6} & 65.2 & 56.5 & 38.6 & 47.3 & 65.2 & 60.9 & \underline{42.4} & 48.4 & 58.7 \\
\midrule
\textbf{DeFA (ours)} & \textbf{72.3} & \textbf{52.7} & \textbf{57.1} & \textbf{71.7} & \textbf{68.5} & \textbf{48.9} & \textbf{55.4} & \underline{69.0} & \textbf{70.7} & \textbf{50.0} & \textbf{56.0} & \underline{69.0} \\
\bottomrule
\end{tabular*}
\end{table}

\subsection{Main results on Who\&When}
\label{sec:whowhen-results}

DeFA achieves the highest Agent and exact Step accuracy with all three backbones (Table~\ref{tab:whowhen-main}). Exact Step improves by 7.6--9.8 percentage points over the strongest baseline per backbone, reaching 52.7\% with GLM-5.2. The accompanying gains in Agent accuracy demonstrate its effectiveness in identifying the responsible agent and the decisive error step.

DeFA also leads in Step~$\pm1$ with every backbone. Although the strongest baseline changes across models, DeFA consistently maintains its precise-localization advantage. These results support connecting step content with its upstream sources and downstream effects as an effective basis for diagnosing multi-agent failures.

\subsection{Main results on Who\&When Pro}
\label{sec:whowhen-pro-results}

\begin{table}[!htb]
\centering
\caption{Failure attribution accuracy on the Who\&When Pro text subset (\%).}
\label{tab:whowhen-pro-main}
\footnotesize
\setlength{\tabcolsep}{1.5pt}
\renewcommand{\arraystretch}{1.0}
\begin{tabular*}{\textwidth}{@{\extracolsep{\fill}}l*{15}{r}@{}}
\toprule
& \multicolumn{5}{c}{GLM-5.2} & \multicolumn{5}{c}{DeepSeek-V4-Pro} & \multicolumn{5}{c}{Qwen3.8-Max} \\
\cmidrule(lr){2-6}\cmidrule(lr){7-11}\cmidrule(lr){12-16}
Method & Agent & Step & $\mathrm{S}_{\pm1}$ & $\mathrm{S}_{\pm3}$ & Mode & Agent & Step & $\mathrm{S}_{\pm1}$ & $\mathrm{S}_{\pm3}$ & Mode & Agent & Step & $\mathrm{S}_{\pm1}$ & $\mathrm{S}_{\pm3}$ & Mode \\
\midrule
All-at-Once & 67.4 & 69.7 & 80.0 & 89.3 & \underline{28.0} & 71.9 & 68.3 & 79.3 & 90.3 & \underline{27.3} & \underline{69.7} & \underline{71.7} & \underline{83.3} & 91.7 & 26.3 \\
Step-by-Step & 66.3 & 69.0 & 77.3 & 87.7 & 27.7 & 69.7 & 66.0 & 74.7 & 86.7 & 24.0 & \textbf{75.3} & \underline{71.7} & 78.3 & 87.0 & \underline{27.0} \\
Binary Search & 60.7 & 64.3 & 74.0 & 91.7 & 25.0 & 56.2 & 68.3 & \textbf{83.7} & 92.7 & 25.0 & 68.5 & 70.3 & 83.0 & 92.0 & 25.7 \\
AgentRx & 66.3 & 68.0 & \underline{81.7} & 89.3 & --- & 67.4 & 69.0 & 81.7 & \underline{93.3} & --- & 59.6 & 63.7 & \textbf{83.7} & 92.3 & --- \\
DeepDebug & 64.0 & 68.7 & 79.3 & 91.0 & --- & 67.4 & 69.0 & 80.0 & 91.7 & --- & 65.2 & 68.0 & 78.3 & 90.0 & --- \\
ErrorProbe & 70.8 & 70.3 & 79.7 & 92.3 & --- & 62.9 & 66.7 & 74.3 & 87.7 & --- & 67.4 & 68.0 & 79.0 & 91.3 & --- \\
CockpitHAT & \underline{77.5} & \underline{72.0} & 81.3 & \underline{93.7} & --- & \underline{74.2} & \underline{72.3} & \underline{82.7} & \textbf{94.0} & --- & 68.5 & 69.3 & 80.7 & \textbf{94.0} & --- \\
\midrule
\textbf{DeFA (ours)} & \textbf{85.4} & \textbf{80.3} & \textbf{89.3} & \textbf{95.3} & \textbf{41.0} & \textbf{79.8} & \textbf{76.3} & 80.0 & 92.7 & \textbf{34.0} & \textbf{75.3} & \textbf{77.7} & \underline{83.3} & \underline{92.7} & \textbf{37.7} \\
\bottomrule
\end{tabular*}
\end{table}

On Pro text, DeFA achieves the highest Agent and exact Step accuracy with all three backbones (Table~\ref{tab:whowhen-pro-main}). Exact Step gains over the strongest baseline range from 4.0 to 8.3 percentage points, reaching 80.3\% with GLM-5.2. This consistent performance on a benchmark spanning eight agent frameworks further supports DeFA's applicability to diverse agent executions.

DeFA also leads taxonomy-aligned methods in Mode accuracy, reaching 34.0--41.0\% and improving over the strongest comparable baseline by 6.7--13.0 percentage points. Together, the Step and Mode results demonstrate its effectiveness in both locating errors and characterizing their nature.

\subsection{Trajectory length and LLM calls}
\label{sec:length-cost}

\begin{figure}[!tbp]
    \centering
    \includegraphics[width=\textwidth]{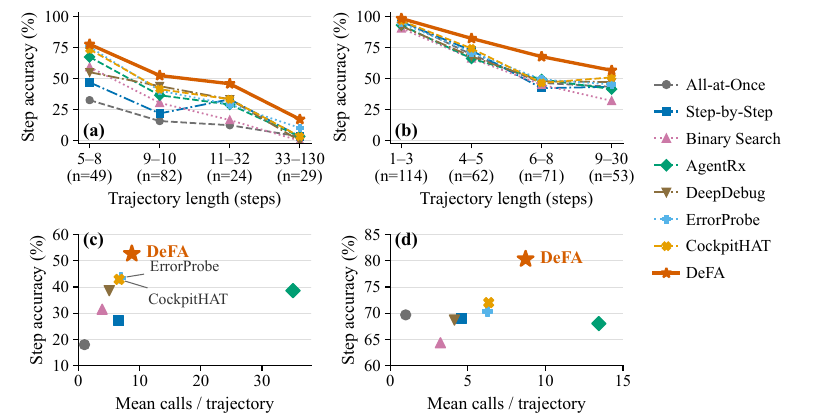}
    \caption{GLM-5.2 exact Step accuracy on Who\&When (left) and Pro text (right), by (a,b) trajectory length and (c,d) mean LLM calls per trajectory.}
    \label{fig:length-cost}
\end{figure}

\paragraph{Localization gains across trajectory lengths.}
Figure~\ref{fig:length-cost}a,b shows exact Step accuracy across trajectory lengths with GLM-5.2. Accuracy generally declines as trajectories grow longer, but DeFA outperforms the baselines in every length interval on both benchmarks. The largest gains occur at intermediate lengths: DeFA reaches 45.8\% on Who\&When trajectories of 11--32 steps, exceeding the strongest baseline by 12.5 percentage points, and improves accuracy by 18.3 points on Pro trajectories of 6--8 steps (67.6\% versus 49.3\%). These results show that DeFA's overall improvement spans different trajectory lengths. Nevertheless, its accuracy remains only 17.2\% in the longest Who\&When interval of 33--130 steps despite leading the baselines, indicating that exact localization in long trajectories remains challenging.

\paragraph{LLM calls and Step accuracy.}
We compare each method's mean number of LLM calls per trajectory with its exact Step accuracy (Figure~\ref{fig:length-cost}c,d). DeFA achieves the highest Step accuracy on both benchmarks. On Who\&When, DeFA reaches 52.7\% with 8.73 calls on average, compared with 43.5\% and 6.93 calls for the strongest baseline, ErrorProbe. On Pro, DeFA reaches 80.3\% with 8.73 calls, compared with 72.0\% and 6.35 calls for the strongest baseline, CockpitHAT.

\subsection{Extension to multimodal trajectories}
\label{sec:multimodal}

To assess DeFA's applicability beyond text-only trajectories, we further evaluate its attribution performance on multimodal trajectories containing image and video evidence. We construct multimodal inputs by combining images or video frames with the text of their corresponding steps, preserving this alignment during context construction so that DeFA can attribute failures using both textual and visual evidence.

We evaluate all eight methods with Qwen3.8-Max and Kimi-K3 on a stratified sample of 70 image and 30 video trajectories from Who\&When Pro. All methods use the same benchmark visual inputs and retain their respective attribution procedures. Step, Step~$\pm1$, Step~$\pm3$, and Mode use all 100 trajectories; following the official Who evaluation protocol, Agent accuracy is computed on the 17 trajectories from multi-agent frameworks.

\begin{table}[!htbp]
\centering
\caption{Multimodal failure attribution accuracy on Who\&When Pro (\%).}
\label{tab:multimodal}
\small
\setlength{\tabcolsep}{2.5pt}
\renewcommand{\arraystretch}{1.0}
\begin{tabular*}{\textwidth}{@{\extracolsep{\fill}}l*{10}{r}@{}}
\toprule
& \multicolumn{5}{c}{Qwen3.8-Max} & \multicolumn{5}{c}{Kimi-K3} \\
\cmidrule(lr){2-6}\cmidrule(lr){7-11}
Method & Agent & Step & $\mathrm{S}_{\pm1}$ & $\mathrm{S}_{\pm3}$ & Mode & Agent & Step & $\mathrm{S}_{\pm1}$ & $\mathrm{S}_{\pm3}$ & Mode \\
\midrule
All-at-Once & 58.8 & 68.0 & 80.0 & 88.0 & \underline{42.0} & \underline{70.6} & 68.0 & 79.0 & 89.0 & \underline{40.0} \\
Step-by-Step & \underline{70.6} & 64.0 & 74.0 & 87.0 & 33.0 & 58.8 & 61.0 & 69.0 & 81.0 & 32.0 \\
Binary Search & \underline{70.6} & 69.0 & 79.0 & 89.0 & 37.0 & 58.8 & 67.0 & 80.0 & \textbf{91.0} & 35.0 \\
AgentRx & 64.7 & 64.0 & 74.0 & 85.0 & --- & 58.8 & 70.0 & \textbf{83.0} & 88.0 & --- \\
DeepDebug & 58.8 & 69.0 & \underline{82.0} & \textbf{92.0} & --- & 64.7 & 65.0 & \underline{82.0} & \textbf{91.0} & --- \\
ErrorProbe & 58.8 & 66.0 & 78.0 & 89.0 & --- & 47.1 & 70.0 & 81.0 & 89.0 & --- \\
CockpitHAT & \underline{70.6} & \underline{73.0} & \underline{82.0} & \underline{91.0} & --- & 64.7 & \underline{72.0} & 81.0 & 87.0 & --- \\
\midrule
\textbf{DeFA (ours)} & \textbf{76.5} & \textbf{80.0} & \textbf{84.0} & 89.0 & \textbf{50.0} & \textbf{82.4} & \textbf{78.0} & \underline{82.0} & \underline{90.0} & \textbf{49.0} \\
\bottomrule
\end{tabular*}
\end{table}

DeFA retains its attribution advantage on multimodal trajectories, leading Agent and exact Step with both backbones and Mode among taxonomy-aligned methods (Table~\ref{tab:multimodal}). Step reaches 80.0\% with Qwen3.8-Max and 78.0\% with Kimi-K3, exceeding CockpitHAT by 7.0 and 6.0 percentage points; Mode reaches 50.0\% and 49.0\%.

These results extend DeFA's effectiveness to executions with visual evidence. Keeping images and video frames aligned with their steps enables dependency and propagation analysis to relate visual observations to subsequent decisions and errors. Gains in both localization and classification support the framework's applicability to multimodal failure diagnosis.

\subsection{Component ablations}
\label{sec:component-ablations}

We evaluate five ablations with GLM-5.2 on Who\&When. Removing segmentation treats each trajectory as one segment; removing the event dependency graph omits dependency context; removing the failure propagation graph skips propagation analysis. A fourth variant removes both graphs. The fifth removes only failure propagation edges and their explanations from final adjudication, retaining all steps and their evidence. All graph variants retain segmentation. Coverage is the fraction of trajectories whose annotated error step is in the final adjudication set $A$.

\begin{table}[!htbp]
\centering
\caption{Component ablations on Who\&When with GLM-5.2 (\%).}
\label{tab:component-ablation}
\small
\setlength{\tabcolsep}{5pt}
\renewcommand{\arraystretch}{1.0}
\begin{tabular*}{\textwidth}{@{\extracolsep{\fill}}lrrrr@{}}
\toprule
Setting & Coverage & Step & Agent & Step $\pm1$ \\
\midrule
Full DeFA & \textbf{83.2} & \textbf{52.7} & \textbf{72.3} & \textbf{57.1} \\
w/o segmentation & 75.0 & 44.6 & 60.9 & 53.3 \\
w/o failure propagation graph & 78.8 & 40.8 & 56.5 & 51.6 \\
w/o event dependency graph & 77.7 & 42.4 & 58.2 & 49.5 \\
w/o both graphs & 73.4 & 36.4 & 53.8 & 46.2 \\
w/o failure propagation edges & \textbf{83.2} & 40.2 & 57.1 & 48.9 \\
\bottomrule
\end{tabular*}
\end{table}

Full DeFA leads all four metrics (Table~\ref{tab:component-ablation}). Removing segmentation lowers Step from 52.7\% to 44.6\%. Without the event dependency graph, Coverage falls to 77.7\% and Step to 42.4\%, supporting dependency context for candidate recall and tracing how steps use earlier information. Removing the failure propagation graph reduces Step to 40.8\%; removing both graphs to 36.4\%. Removing only propagation edges retains the same adjudication set and 83.2\% Coverage, but lowers Step to 40.2\%, a 12.5-point drop. This supports using propagation relations to trace how errors are introduced, transmitted, and adopted when locating the decisive error.

\subsection{Downstream utility: skill improvement}
\label{sec:downstream-skills}

We integrate DeFA into Trace2Skill~\citep{ni2026trace2skill}. Trace2Skill analyzes agent execution trajectories to identify the causes of failure in failed trajectories and patterns in successful trajectories, using these analyses to evolve reusable skills. We replace the failed-trajectory analysis module in its Combined setting with DeFA.

\paragraph{Experimental setup.}
We use 100 SpreadsheetBench-Verified~\citep{ma2024spreadsheetbench} tasks for skill evolution and another 100 for testing. We evaluate Kimi-K2.5 and Qwen3.5-Plus, each with a human-written \texttt{xlsx} skill or a model-generated initial skill, yielding four settings. Following the original Trace2Skill setup, skill evolution and testing use the same model.

\begin{figure}[!htbp]
\centering
\includegraphics[width=\textwidth]{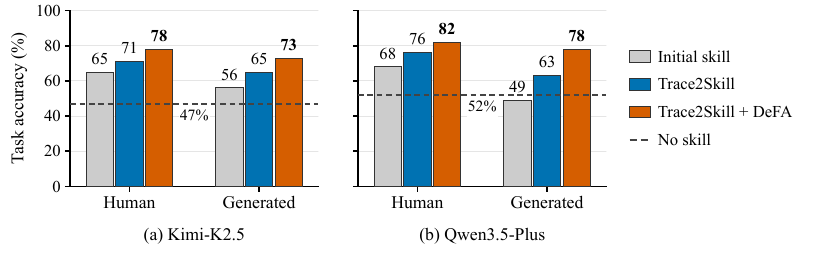}
\caption{Test accuracy after skill evolution on 100 SpreadsheetBench-Verified tasks. Human and Generated denote human-written and model-generated initial skills; dashed lines denote no skill.}
\label{fig:trace2skill-downstream}
\end{figure}

\paragraph{Effect on skill evolution.}
With DeFA feedback, test accuracy improves over native Trace2Skill in all four settings by 6--15 percentage points, averaging 9.0 points (Figure~\ref{fig:trace2skill-downstream}). The largest gain occurs with Qwen3.5-Plus and a model-generated initial skill, where accuracy rises from 63\% to 78\%. Both models benefit under each initialization, showing that DeFA's diagnostic feedback can improve human-written and model-generated skills for subsequent tasks. Trajectory diagnosis and its full analytical outputs may also provide feedback for broader agent self-improvement and continual evolution.

\section{Conclusion and Limitations}
\label{sec:conclusion}

We introduced DeFA, which traces agent failures through step dependencies. It combines protocol relations and semantic dependencies into an event dependency graph spanning the trajectory to capture how steps use earlier information. DeFA then identifies events that may violate task requirements and traces their sources and effects to construct a failure propagation graph. Final attribution uses this connected evidence to compare how steps introduce, transmit, or adopt errors and identify the decisive error and responsible agent.

Experiments on Who\&When and Who\&When Pro demonstrate precise attribution gains across backbones, trajectory lengths, and text and multimodal inputs. Ablations support segmentation, the event dependency graph, and the failure propagation graph. DeFA's diagnostic feedback also improves test accuracy after Trace2Skill skill evolution by an average of 9.0 percentage points over the native pipeline, showing that the diagnoses can guide skill improvements for subsequent tasks.

\paragraph{Limitations.}
DeFA's semantic dependency recovery and failure propagation analysis depend on the model's trajectory interpretation; missing or misidentified relations can affect attribution, especially in long executions with dispersed evidence. Downstream utility has been evaluated for spreadsheet skill evolution; effectiveness in other domains and forms of agent improvement remains to be assessed. Future work can improve relation identification in long trajectories and broaden the applications of diagnostic feedback.

\clearpage
\subsection*{AI use statement}

We used generative AI tools to assist with manuscript editing and translation, and with writing and revising parts of the research code. All AI-assisted text and code were reviewed and checked by the authors. The authors take full responsibility for the final manuscript, implementation, and reported results.

\bibliography{references}
\bibliographystyle{iclr2027_conference}

\end{document}